\documentclass{applemlr}
\usepackage{amsmath}
\usepackage{enumerate}
\usepackage{algorithm}
\usepackage{algpseudocode}
\usepackage{amsfonts}
\usepackage{amsthm}
\usepackage{cleveref}
\usepackage{diagbox}
\usepackage{colortbl}
\usepackage{amssymb}
\usepackage{xspace}
\usepackage{wrapfig}
\usepackage{adjustbox}
\usepackage{tabularx}
\usepackage{booktabs}
\usepackage{mathtools}
\usepackage{tikz}
\usepackage{enumitem}
\usepackage{silence}
\usepackage{dsfont}
\usepackage[table]{xcolor}
\usepackage[dvipsnames]{xcolor}
\usepackage{multirow}
\usepackage{makecell}
\usepackage{xfakebold}

\usepackage{amsmath,amsfonts,bm}

\def\eqref#1{equation~\ref{#1}}

\def\1{\bm{1}}

\DeclareMathAlphabet{\mathsfit}{\encodingdefault}{\sfdefault}{m}{sl}
\SetMathAlphabet{\mathsfit}{bold}{\encodingdefault}{\sfdefault}{bx}{n}

\definecolor{textgray}{HTML}{6E6E73}
\usetikzlibrary{positioning, calc}
\usetikzlibrary{decorations.pathmorphing}

\makeatletter
\patchcmd{\wrong@fontshape}{\@gobbletwo}{}{}{}
\makeatother
\numberwithin{equation}{section}
\makeatletter
\AtBeginDocument{
  \urlstyle{sf}
  
}
\makeatother

\definecolor{light}{RGB}{125, 125, 125}
\crefname{tcb@cnt@pbox}{code}{code}
\Crefname{tcb@cnt@pbox}{Code}{Code}
\crefname{assumption}{assumption}{assumption}
\Crefname{assumption}{Assumption}{Assumptions}

\newtcolorbox[auto counter]{pbox}[2][]{
  colback=white,
  title=Code~\thetcbcounter: #2,
  #1,fonttitle=\sffamily,
  fontupper=\sffamily,
  arc=2pt,
  colframe=bgcolor,
  coltitle=fgcolor,
  colbacktitle=bgcolor,
  toptitle=0.25cm,
  bottomtitle=0.125cm
}

\makeatletter
\newcommand\applefootnote[1]{%
  \begingroup
  \renewcommand\thefootnote{}%
  \renewcommand\@makefntext[1]{\noindent##1}%
  \footnote{#1}%
  \addtocounter{footnote}{-1}%
  \endgroup
}
\makeatother

\definecolor{cverbbg}{gray}{0.90}

\usepackage{graphicx}
\usepackage{colortbl}
\usepackage{multirow}
\usepackage{multicol}
\usepackage{bm}
\usepackage{subcaption}
\usepackage{diagbox}
\usepackage{amsmath,mathtools}
\usepackage{amssymb}
\usepackage{animate}
\usepackage{setspace}
\usepackage{booktabs} 
\usepackage{tcolorbox}

\usepackage{tabularx}
\usepackage{xspace}
\usepackage{interval}
\usepackage{siunitx}
\usepackage{epigraph}
\usepackage{inconsolata}
\usepackage{caption}
\usepackage{float}
\usepackage{soul}
\usepackage{pifont}
\usepackage{makecell} 
\usepackage{wrapfig}
\usepackage{natbib}

\newcounter{prompt}
\newcommand{\prompt}[4]{%
\refstepcounter{prompt}%
\begin{tcolorbox}[
    float=#1,
    colback=lightblue!35,
    colframe=white!45!black,
    title={Prompt.~\theprompt:~#2}
]
#3
\label{#4}
\end{tcolorbox}
}

\renewcommand{\thefootnote}{\fnsymbol{footnote}}

\definecolor{cvprblue}{rgb}{0.21,0.49,0.74}
\crefname{prompt}{Prompt.}{Prompts.}

\definecolor{lightblue}{rgb}{0.93, 0.96, 1.0}
\definecolor{lightgray}{rgb}{0.92, 0.92, 0.92}
\definecolor{lightgreen}{rgb}{0.89, 0.9375, 0.90625}

\newcommand{\systemold}{\textit{UniGen}\xspace}
\newcommand{\system}{\textit{UniGen-1.5}\xspace}
\newcommand{\ie}{i.e.}
\newcommand{\eg}{e.g.}
\title{\system: Enhancing Image Generation and Editing through Reward Unification in Reinforcement Learning}

\author{Rui Tian$^{1\,2 \,*}$, \,Mingfei Gao$^{2 \,\dagger \,\ddagger}$, \,Haiming Gang$^2$, \,Jiasen Lu$^2$, \,Zhe Gan$^2$,  \,Yinfei Yang$^2$, \,Zuxuan Wu$^{1\,\ddagger}$, \,\\ Afshin Dehghan$^{2}$
}

\vspace{-2pt}
\affiliation{$^1$Institute of Trustworthy Embodied AI, Fudan University \quad $^2$Apple \\ 
$^\dagger$Project lead; \,$^\ddagger$Corresponding authors}

\abstract{
We present \system, a unified multimodal large language model (MLLM) for advanced image understanding, generation and editing. Building upon \systemold, we comprehensively enhance the model architecture and training pipeline to strengthen the image understanding and generation capabilities while unlocking strong image editing ability. Especially, we propose a unified Reinforcement Learning (RL) strategy that improves both image generation and image editing jointly via shared reward models. 
To further enhance image editing performance, we propose a light Edit Instruction Alignment stage that significantly improves the editing instruction comprehension that is essential for the success of the RL training. Experimental results show that \system demonstrates competitive understanding and generation performance. Specifically, \system achieves 0.89 and 4.31 overall scores on GenEval and ImgEdit that surpass the state-of-the-art models such as BAGEL and reaching performance comparable to proprietary models such as GPT-Image-1.
}  
\date{\sffamily\today}

\begin{document}
\maketitle
\footnotetext[1]{Work done while at Apple.}

\begin{center}
    \centering
    \vspace{-0.15cm}
    \begin{minipage}{\linewidth}
        \setstretch{0.6}
        \includegraphics[width=0.95\linewidth]{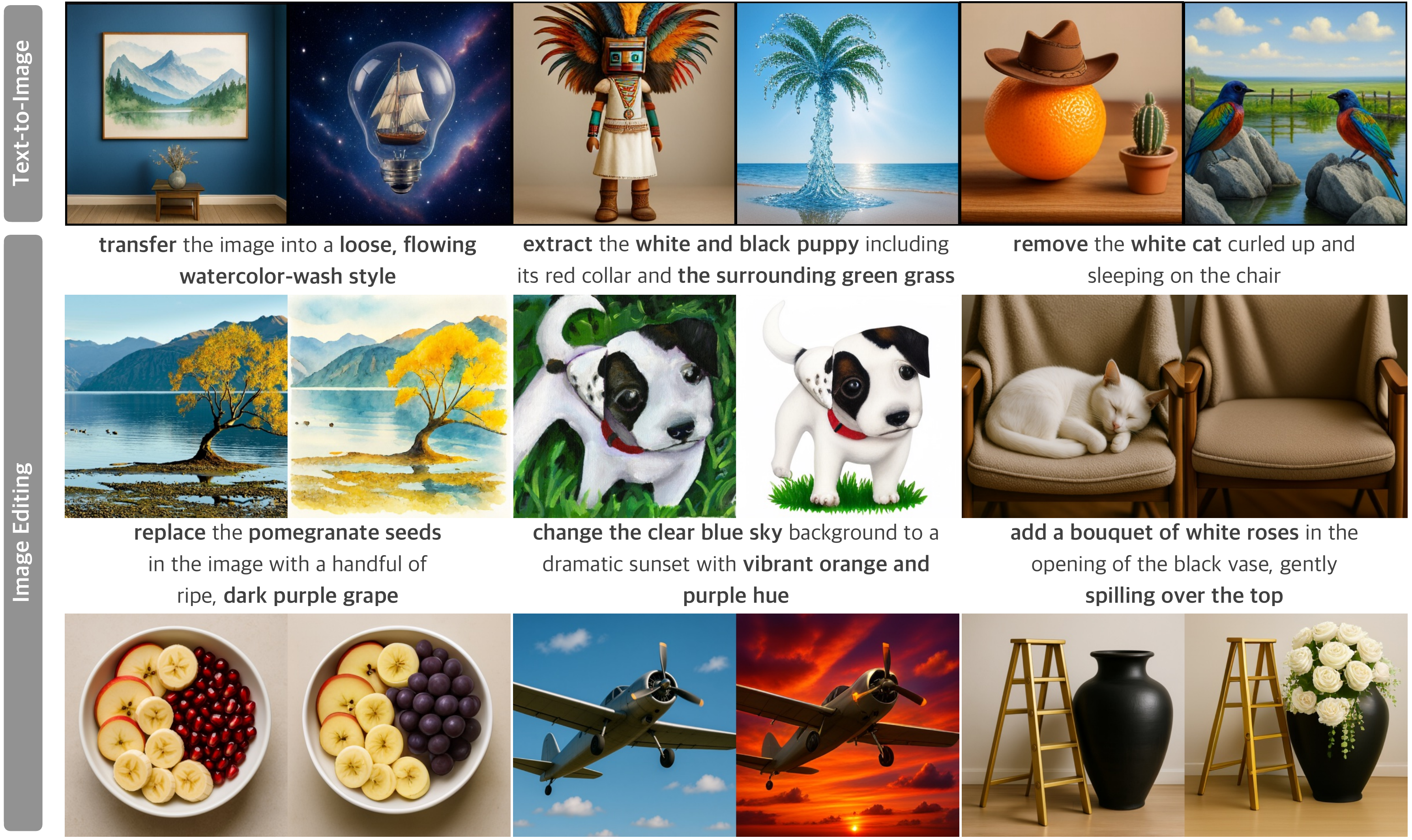}\\
        \centering
    \end{minipage}
    \vspace{-15pt}
    \captionof{figure}{Examples of images generated by \system.
    }
    \label{fig:cover}
\end{center}

\section{Introduction}
\label{sec:intro}

Unified Multimodal Large Language Models (MLLMs)~\cite{chen2025blip3,deng2025bagel,lin2025uniworld,xie2025show,jiao2025unitoken,ai2025ming,li2025manzano} have achieved promising performance across both visual understanding and generation domains. Taking the advantage of strong reasoning ability and knowledge-rich representations of Large Language Models (LLMs)~\cite{zhang2024mm1,mckinzie2024mm1,yang2024qwen2,yang2025qwen3}, unified MLLMs have demonstrated superiority in achieving better semantic consistency compared to vanilla image-generation-only models~\cite{betker2023improving,han2024infinity,podellsdxl} that rely on conditions from text encoders. 

Among recent unified MLLMs, \systemold~\cite{tian2025unigen} introduced an effective data-centric pipeline for building a competitive model from pre-training to post-training stages. Specifically in post-training, it leverages its intrinsic understanding capability to enhance its generation performance via a chain-of-thought verification (CoT-V) strategy. While CoT-V effectively improves performance on text-to-image generation, it also introduces substantial inference overheads. Additionally, \systemold lacks the ability of image editing~\cite{wu2025omnigen2,xiao2025omnigen,liu2025step1x} that is considered as the core for measuring fine-grained controllability of content generation.

We introduce \system, that significantly improves \systemold with a focus on the post-training stages. We design an effective model architecture for \system that supports image understanding, generation as well as editing within a single model. Moreover, we observe that model remains inadequate in handling diverse editing scenarios after supervised fine-tuning due to its insufficient comprehension of the editing instructions. Therefore, we propose \emph{Edit Instruction Alignment} as a light Post-SFT stage to enhance the alignment between editing instruction and the semantics of the target image. Specifically, it takes the condition image and the instruction as inputs and is optimized for predicting the semantic content of the target image via textual descriptions. Experimental results suggest that this stage is highly beneficial for boosting the editing performance.

RL has demonstrated great potential for improving text-to-image generation~\cite{jiang2025t2i,wang2025simplear,liu2025flow,guo2025can} by encouraging path exploration without incurring huge inference overheads compared to test-time scaling methods such as CoT-V in~\cite{tian2025unigen}. However, fewer works~\cite{wei2025skywork} have demonstrated effective ways for elevating image editing using RL for unified MLLMs. We propose that image editing is a more challenging task that involves complicated variations ranging from very subtle changes such as removing/replacing small objects to substantial changes such as altering image style in the pixel space. This raises a big challenge for robust reward modeling. To relieve the issue, we propose to reformulate the image editing task as a general image generation task and optimize it together with the standard text-to-image task via shared reward models under the same schema of RL. 
Similar to text-to-image generation, we supervise image editing training using the RL reward signals built from directly measuring the alignment between the generated/edited image and its text description. This strategy unlocks us to use the stable text-to-image reward models~\cite{wu2023human,xu2023imagereward,wang2025unified} for jointly improving both tasks. 

Through the efforts above, \system provides a stronger baseline for advancing research on unified MLLMs and establishes competitive performance across image understanding, generation, and editing benchmarks. The experimental results show that \system obtains 0.89 and 86.83 on GenEval and DPG-Bench, significantly outperforming recent methods such as BAGEL~\cite{deng2025bagel} and BLIP3o~\cite{chen2025blip3}.
For image editing, \system achieves 4.31 overall scores on ImgEdit, surpassing recent open-sourced models such as OminiGen2~\cite{wu2025omnigen2} and is comparable to proprietary models such as GPT-Image-1.

Our contributions are summarized as follows:
\begin{itemize}[leftmargin=20pt]
    \item We present \system, a unified MLLM equipped with an effective model architecture and training pipeline for advanced image understanding, generation and editing.

    \item We design a unified RL training schema that optimizes image editing and generation using shared reward models that significantly boosts the performance of both tasks.

    \item We propose \emph{Edit Instruction Alignment} as a Post-SFT stage that significantly improves the editing performance via enhancing edit instruction comprehension.

    \item \system achieves competitive performance against state-of-the-art unified MLLMs. As shown in \Cref{fig:cover}, we obtain competitive performance on image editing (on-par with GPT-Image-1 on ImgEdit benchmark) and image generation (significantly outperforming BLIP3o on GenEval and DPG-Bench). We also attain strong results on image understanding (comparable to Show-o2~\cite{xie2025show}).
\end{itemize}

\begin{figure*}[ht]
    \centering
    \includegraphics[width=\linewidth]{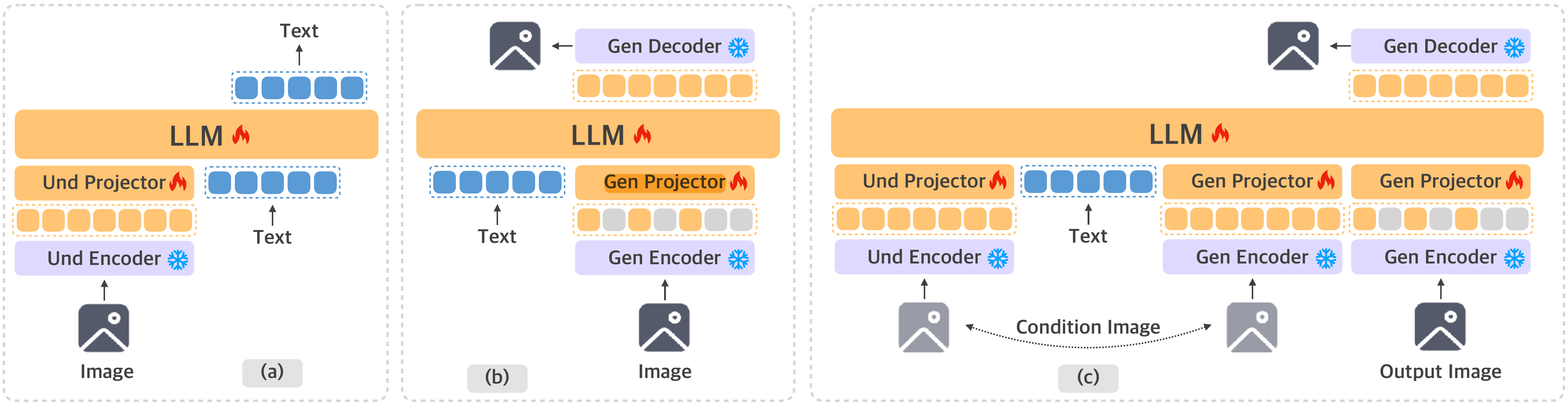}
    \caption{
        \textbf{The architecture of \system} jointly optimized for \textbf{(a)} image understanding, \textbf{(b)} text-to-image generation and \textbf{(c)} image editing. See more details in \Cref{sec:method_arch}.
    }
    \label{fig:architecture}
\end{figure*}

\section{Related Work}
There is a growing research trend towards building unified multimodal models capable of image understanding and generation within a single model or framework. Existing approaches can be broadly categorized into three paradigms. First, the unified autoregressive (AR) approach encodes images as either discrete tokens~\cite{team2024chameleon,wang2024emu3,wu2024vila,chen2025janus,geng2025x} or continuous visual embeddings~\cite{sun2023emu,sun2024generative,ge2024seed,tong2025metamorph,fan2025unified}, allowing LLMs to treat vision and text as a unified sequence for joint autoregressive prediction. Second, the decoupled LLM-diffusion approach~\cite{pan2025transfer,wu2025qwen,wu2025omnigen2,chen2025blip3} separates reasoning from image generation, using a frozen LLM for multimodal understanding while offloading image synthesis to a diffusion-based decoder. Third, the hybrid AR-diffusion approach~\cite{deng2025bagel,zhou2024transfusion,liang2024mixture,xie2024show} integrates both the AR and diffusion paradigms within a single transformer that autoregressively generates text while employing embedded diffusion for visual output. Orthogonal to these modeling strategies, visual tokenizers are central in supporting both semantic understanding and high-fidelity generation. Recent studies have explored both decoupled encoders~\cite{wu2024janus} and unified tokenizers~\cite{jiao2025unitoken,li2025manzano,ma2025unitok} to achieve better task balancing.  Additionally, emerging research investigates the use of reinforcement learning (RL) to enhance native image generation quality~\cite{geng2025x,chen2025blip3o,wei2025skywork,mao2025unirl,jiang2025co}, which is also the focus of our work. Building upon the \systemold framework~\cite{tian2025unigen} which uses masked token prediction for image generation, we effectively integrates image generation and editing within a single RL training framework and optimize for both tasks via shared reward models.

\section{Methods}
\label{sec:methods}

\subsection{Architecture}
\label{sec:method_arch}

We build \system upon a pre-trained LLM, \ie, Qwen2.5-7B~\cite{yang2025qwen3}, and leverage separate encoders for understanding and generation. As shown in ~\Cref{fig:architecture}, we use the discrete visual tokenizer (MAGViTv2~\cite{yulanguage}) for visual generation and continuous visual encoder (SigLIP2~\cite{tschannen2025siglip}) for visual understanding.

\vspace{2pt}
\noindent \textbf{For image understanding}, we utilize SigLIP2 as our visual encoder $\mathbf{Enc}^{U}$. Comparing to SigLIP with fixed input resolution, e.g., $384\times384$, SigLIP2 can receive images with varying input sizes of arbitrary aspect ratio that is important for maintaining images' native information. An input image $X^{U}$ will be projected to a set of continuous tokens $\mathcal{X}^{U}=\mathbf{Enc}^{U}(X^{U})$ dependent on its original size. Following the LLaVA~\cite{liu2023visual} workflow, an MLP-based projector is adopted to align the image and text embeddings into the same space and then the visual embedding together with text embedding are fed into the LLM for response generation via next-token prediction as shown in \Cref{fig:architecture} (a).

\vspace{2pt}
\noindent \textbf{For text-to-image generation}, we generally adopt the same setting as \systemold by using masked token prediction~\cite{chang2022maskgit} as our training objective. For each image $X^{G}$, we encode it into a sequence of discrete tokens with the generation tokenizer $\mathcal{X}^{G}=\mathbf{Enc}^{G}(X^G)$. The model is trained to generate target image tokens conditioned on a text prompt $\mathcal{T}_C$. During training, we randomly sample a binary mask $\in \{0, 1\}$ for each token, given a masking ratio $\eta$ according to a masking scheduling function $\gamma(\cdot)$.
For each token with mask equal to $1$, we replace
its corresponding discrete image token $\mathcal{X}^{G}_i$ with a special mask token $\mathtt{[MASK]}$  to form the final input image sequence. As shown in \Cref{fig:architecture} (b), the LLM takes the text prompt and the masked image sequence tokens as inputs and optimizes for predicting the masked visual tokens back. During inference, the image generation starts with all masked tokens and perform masked token prediction in multiple turns. We set the image-generation resolution to $384 \times 384$.

\vspace{2pt}
\noindent \textbf{For image editing}, we unlock this capability during the supervised fine-tuning stage. Given a condition image $X_{C}$, and an editing text prompt $\mathcal{T}_C$, we leverage both the understanding encoder and the generation tokenizer, to obtain $\mathcal{X}_{C}^{U}=\mathbf{Enc}^{U}(X_{C})$ and $\mathcal{X}_{C}^{G}=\mathbf{Enc}^{G}(X_{C})$ that extract the continuous (semantic) and discrete (low-level) features from the condition image. We resize condition image to 384$\times$384 for feature extraction to ensure capturing substantial details. After projecting the features into a joint space via MLP layers, we sequentially concatenate the semantic visual embedding, the text embedding and the low-level visual embedding (see \Cref{fig:architecture} (c)). We then feed the assembled sequence as the condition for image editing into LLM. 
The goal is to generate discrete visual tokens $\mathcal{X}_{O}^{G}$ of the output image $\mathcal{X}_{O}$, where $\mathcal{X}_{O}^{G}=\mathbf{Enc}^{U}(X_{O})$. Similar to the text-to-image generation, we utilize the masked token prediction strategy for image token prediction. The generation resolution for editing is set to $384\times384$.

\subsection{Pre-training}
\label{sec:pretrain}
During the pre-training stage of \system, we aim to develop foundational visual captioning and generation capabilities with a large collection of well-aligned image–text pairs. Specifically, we employ the pre-training data with fine-grained captions from \systemold, composed of ImageNet~\cite{ridnik2021imagenet} , CC-3M~\cite{sharma2018conceptual}, CC-12M~\cite{changpinyo2021conceptual} and SAM-11M~\cite{kirillov2023segment}. We also include a small portion of text-only training data from RefinedWeb~\cite{penedo2023refinedweb} to maintain LLM's basic language ability. For simplicity, we adopt only one pre-training stage and unfreeze all the parameters except for the $\mathbf{Enc}^{U}$ and $\mathbf{Enc}^{G}$. We include image understanding and text-to-image generation tasks in this stage and set both the resolution of image inputs for generation and understanding to $384\times384$. We construct each training batch by sampling data from image generation, image understanding and text understanding with a ratio of 3:2:1.

\begin{figure}[t]
    \centering
    \includegraphics[width=0.8\linewidth]{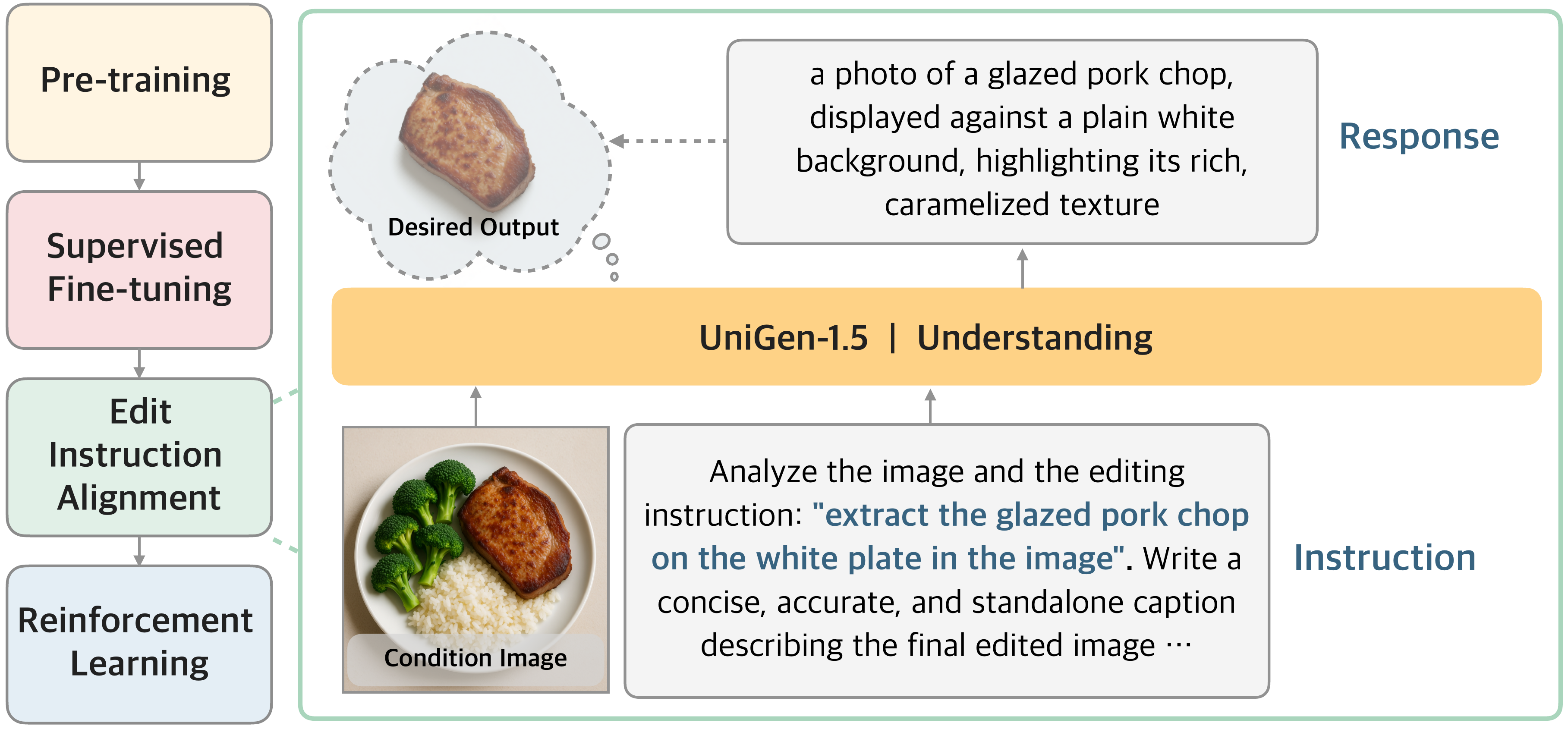}
    \caption{
         Illustration of \emph{Edit Instruction Alignment} in the entire training pipeline of \system. }
    \label{fig:edit_inst_align}
\end{figure}

\subsection{Supervised Fine-tuning}
\label{sec:sft}
In supervised fine-tuning (SFT), we seek to push forward the generation and understanding performance of \system with stronger data mixtures and incentivate \system's image editing ability with joint training.

\vspace{2pt}
\noindent \textbf{Image Generation and Editing.} We follow the architecture introduced in \Cref{sec:method_arch} for both image generation and editing. Inspired by works~\cite{chen2025sharegpt,chen2025blip3,ye2025echo} highlighting the advantage of synthetic data generated by GPT-4o~\cite{gpt4o}, we expand our training mixture by adding the high-quality samples proposed in BLIP-3o~\cite{chen2025blip3} and ShareGPT-4o-Image~\cite{chen2025sharegpt}. Meanwhile, we unlock image editing by enriching our mixture with image editing data sourced from ShareGPT-4o-Image and GPT-Image-Edit-1.5M~\cite{wang2025gpt}. 

\vspace{2pt}
\noindent \textbf{Image Understanding.} We employ the image mixture from SlowFast-LLaVA-1.5~\cite{xu2025slowfast,xu2024slowfast} to enhance the instruction following capability for image understanding. To encourage the model to perceive finer details of the input image while maintaining the training efficiency, we resize an input image according to the following rules: (1) its width and height must be a multiple of $16$ to ensure the compatibility with the patch size of the encoder, (2) the resized image has the closest aspect ratio as its original one, and (3) we maximize the input resolution under the constraint that the number of visual tokens $\leq 2,304$. That is approximate to the number of tokens extracted from an image of $768\times 768$.

\vspace{2pt}
\noindent \textbf{Joint SFT Training.} Similar to the pre-training stage, we optimize for three tasks in each training step including generation (either text-to-image generation or image editing), image understanding and text understanding. We use a ratio of 3:4:1 training samples from the above three tasks. In practice, we apply round-robin sampling of text-to-image generation and image editing in every other training batches to improve the training stability. After this joint SFT training, \system exhibits the new image editing capability. 

\subsection{Editing Instruction Alignment}
\label{sec:edit_inst_align}

During preliminary experiments of RL, we found that for challenging editing instructions, our model often produced candidates that all failed to satisfy the instruction, resulting in small standard deviations in rewards. Under these circumstances, GRPO receives weak learning signals and struggles to effectively improve the policy. We attribute this challenge to the model’s insufficient ability to comprehend complicated editing instructions, therefore not able to accurately infer the semantic content of output images.

To mitigate this issue,  we include \emph{Editing Instruction Alignment} as a Post-SFT stage to enhance the alignment of editing instruction and semantic content of the desired output. As shown in \Cref{fig:edit_inst_align}, \system takes condition image and editing instruction as inputs and is optimized for predicting the textual description of expected output image that forms a vital bridge to the final visual generation. Consequently, the model develops a more faithful understanding of editing intentions, enabling semantically coherent yet diverse candidate generation and providing informative learning signals during RL. See details of training data in \Cref{sec:data_generation_details}.

\begin{figure*}[t]
    \centering
    \includegraphics[width=\linewidth]{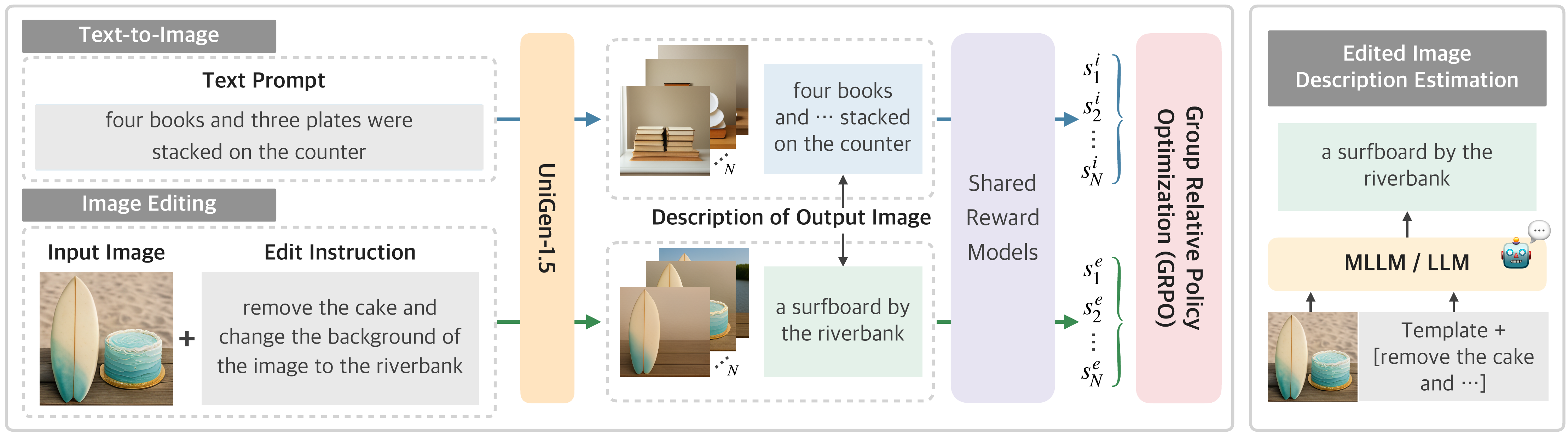}
    \caption{
        \textbf{\textit{Left}: The pipeline of GRPO training in \system}. We utilize shared reward models for both text-to-image generation and image editing. For the former, we directly input the generated image with the {\fboxsep=1pt\colorbox{lightblue}{\textit{text prompt}}} to obtain rewards. For the latter, we get reward signals by measuring the alignment between the {\fboxsep=1pt\colorbox{lightgreen}{\textit{edited image description}}} and the generated image. \textbf{\textit{Right}: The pipeline of edited image description estimation}. We leverage powerful external MLLMs and LLMs to generate the description of desired edited images.}

    \label{fig:unigen_grpo}
\end{figure*}

\subsection{Reinforcement Learning}
\label{sec:grpo}
We improve the overall visual generation quality of \system via a RL stage empowered with group relative policy optimization (GRPO)~\cite{shao2024deepseekmath,guo2025deepseek}. While a series of works highlight the effectiveness of GRPO over improving the performance of text-to-image generation~\cite{wang2025simplear,jiang2025t2i,huang2025interleaving}, its impacts on more generalized forms of visual generation, \eg, image editing, remains underexplored. In \system, we propose to unify the RL training for both text-to-image generation and image editing as shown in \Cref{fig:unigen_grpo}. Specifically, we propose that the output image's quality from both tasks can be assessed by measuring the semantic alignment between the image and its corresponding text description. 

\vspace{2pt}
\noindent\textbf{RL formulation.}
Initialized from our Post-SFT model, \system acts as a policy model, $\pi_{\theta}$, that takes different conditions as inputs and generates the corresponding sequence of visual tokens $\hat{\mathcal{X}}_{O}^G$. For the text-to-image task, the condition is simply the text embedding of the prompt $\mathcal{T}_C$, while for editing task, \system conditions the image generation on $\mathcal{X}_C^U$, an editing text embedding $\mathcal{T}_C$ and $\mathcal{X}_C^G$. During training, we sample $N$ sequences $\{\hat{\mathcal{X}}_{O1}^G, \cdots, \hat{\mathcal{X}}_{ON}^G \}$ as output candidates from $\pi_{\theta}$, each of which will be assigned a scalar reward $R_i$. The rewards are used to calculate a group-normalized advantage as in Eq.\ref{eq:grpo_advantage}.

\begin{equation}
    \label{eq:grpo_advantage}
    A_i = \frac{R_i-mean\{R_1,R_2,...,R_N\}}{std\{R_1,R_2,...,R_N\}}
\end{equation}

The parameters of our policy model are updated by optimizing the training objective in Eq.\ref{eq:grpo_objective}, where $\pi_{ref}$ indicates reference policy (initial policy), $\rho_i=\frac{\pi_{\theta}}{\pi_{\theta_{old}}}$ refers to the importance sampling ratio and $\pi_{\theta_{old}}$ indicates the old policy before update.
\begin{equation}
\label{eq:grpo_objective}
\begin{split}
\mathcal{J}(\theta) &= 
\frac{1}{N}\sum_{i=1}^N 
\min\!\big(\rho_i A_i, \operatorname{clip}(\rho_i, 1-\epsilon, 1+\epsilon) A_i\big) \\
&\quad - \beta\, D_{\mathrm{KL}}(\pi_{\theta} \| \pi_{\mathrm{ref}})
\end{split}
\end{equation}

\noindent \textbf{Shared reward models.} Designing editing rewards is inherently more challenging than designing rewards for text-to-image generation, as visual edits are often subtle, localized, and highly context-dependent.
Moreover, training editing-specific reward models~\cite{luo2025editscore,li2024instructrl4pix} requires substantial manual annotation costs for collecting large-scale image editing data from diverse categories and obtaining high-quality labels that align with human preference. These challenges make it extremely difficult to construct reliable editing rewards at scale.
To this end, we propose to leverage robust, well-developed text-to-image reward models to evaluate edited image. 

Specifically, We introduce a unified RL formulation of image generation and editing by assessing the quality for both tasks using $\mathbf{R}(\tilde{\mathcal{X}}_{O}^G, \mathcal{T_O})$, where $\mathbf{R}(\cdot)$ denotes the shared reward functions, $\tilde{\mathcal{X}}_{O}^G$ indicates the generated image in pixel space and $\mathcal{T_O}$ refers to the text description of expected output. We directly use the ground-truth text prompt as $\mathcal{T_O}$ for text-to-image generation and use the textual caption synthesized by Qwen2.5-72B for image editing (see \Cref{sec:data_generation_details} for details). 
We believe that a powerful LLM is capable of reliably reflecting the visual differences, capturing the details and layout of the edited image in its description, regardless of varying modification magnitude.
Inspired by T2I-R1~\cite{jiang2025t2i}, we opt to implement $\mathbf{R}(\cdot)$ with an ensemble of diverse vision experts to assign rewards for image candidates. Our reward models include CLIP-H~\cite{fang2023data,cherti2023reproducible}, HPSv2~\cite{wu2023human}, Unified-Reward-7B~\cite{wang2025unified} and ORM~\cite{jiang2025t2i}.

\vspace{2pt}
\noindent \textbf{RL Training Data.} For text-to-image generation, we use the training set from T2I-R1~\cite{jiang2025t2i}, including 6,786 prompts sourced from T2I-CompBench~\cite{huang2023t2i} and PARM~\cite{guo2025can}.
For training samples of image editing, we collect the Edit-RL dataset with 10,568 samples. We generate the condition images using Qwen-Image~\cite{wu2025qwen}, and construct versatile edit instructions with Qwen-2.5-VL-72B-instruct~\cite{bai2025qwen2} based on our designed templates. Moreover, we use Qwen-2.5-72B-instruct~\cite{yang2024qwen2} to synthesize output descriptions for the desired images. These pseudo labeled descriptions will be used to calculate the rewards of edited images during training. See details in \Cref{sec:data_generation_details}.

\section{Experiments}
\label{sec:experiment}

\subsection{Implementation Details}
\label{sec:implementation_details}
\vspace{-3pt}

We initialize \system with pre-trained Qwen2.5-7B~\cite{yang2024qwen2} LLM, and adopt MAGVITv2 from Show-o~\cite{xie2024show} as our discrete visual encoder with input resolution of $384\times384$
and siglip2-so400m-patch16-naflex~\cite{tschannen2025siglip} as our continuous visual encoder for native image resolution. For both image generation and editing, we leverage MAGVITv2's decoder to project the visual tokens back to the image pixel space.
Both discrete and continuous encoders are kept frozen across all the training stages.

During pre-training, we use 96 H100-80G GPUs, set the batch size to 576 and adopt the learning rate of $1e^{-4}$. For supervised fine-tuning, we use 64 H100-80G GPUs, set the batch size to 128 and the learning rate to $2e^{-5}$. For the \emph{Edit Instruction Alignment} stage, we train on the collected Edit-Align dataset (See details in \Cref{sec:data_generation_details}) for 500 steps on 8 H100-80G GPUs with a batch size of 64. In this stage, we set the learning rate to $1e^{-5}$ and adopt the cosine schedule. To accommodate the classifier-free guidance during inference, we randomly drop text prompts during training for both text-to-image and image editing tasks with a probability of 10\%, while dropping $\mathcal{X}_C^U$ and $\mathcal{X}_C^G$ for image editing training samples with probabilities of 50\% and 10\%, respectively.
\begin{figure*}[t]
    \centering
    \includegraphics[width=\linewidth]{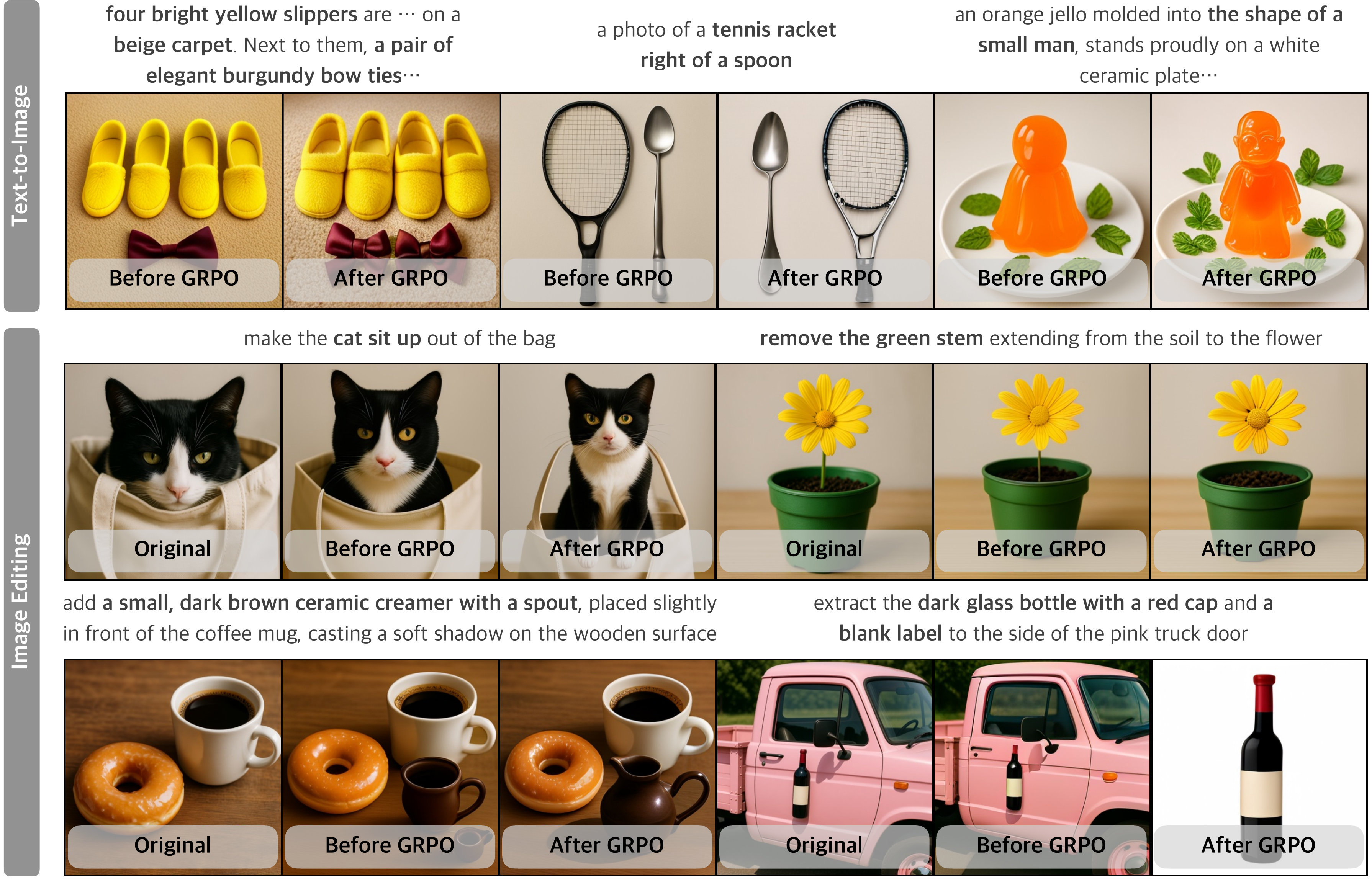}
    \caption{Examples generated by \system, highlighting the contribution of GRPO training.}
    \label{fig:grpo_impact}
\end{figure*}

During GRPO, we follow T2I-R1~\cite{jiang2025t2i} to remove the traditional ratio clipping and simply leverage an explicit KL-penalty regularization to constrain policy updates. We conduct GRPO training for 1500 steps with the learning rate set to  $3e^{-6}$ and the batch size set to 32 on 8 B200 GPUs. We set the KL-penalty coefficient $\beta$ to 0.01 and generate $N=8$ image candidates for each input. To accelerate training with minimal impact on performance, we sample each image candidate using only 16 decoding steps and disable the classifier-free guidance.

\begingroup
    \begin{table*}[t]
    \caption{
        Comparison with baseline models on ImgEdit benchmark. The best and second-best results are highlighted in \textbf{bold} and \underline{underlined}, respectively. \system achieves the best overall score against all the other models.
    }
    \centering
    \footnotesize
    \addtolength{\tabcolsep}{-3pt} 
    \resizebox{\linewidth}{!}{
    \begin{tabular}{lc|cccccccccc}
    \toprule
    \textbf{Model} & \textbf{\#Params} &\textbf{Add} &\textbf{Adjust} &\textbf{Extract} &\textbf{Replace} &\textbf{Remove} & \textbf{Background} & \textbf{Style}& \textbf{Compose}& \textbf{Action}&  \textbf{Overall} \\
    \cmidrule{1-2}\cmidrule{3-12}

    Step1X-Edit~\cite{liu2025step1x}  & 19B &3.88 &3.14 &1.76 &3.40 &2.41 &3.16 &4.63 &2.64 &2.52 &3.06\\
    
    UniWorld-V1 ~\cite{lin2025uniworld} & 7B + 12B & 3.82 &3.64 &2.27 &3.47 &3.24 &2.99 &4.21 &2.96 &2.74 &3.26\\
    BAGEL~\cite{deng2025bagel}  & 7B MoT &3.56 & 3.31 &1.70 &3.30 &2.62 &3.24 &4.49 &2.38 &4.17 &3.20\\
    OmniGen~\cite{xiao2025omnigen}  & 7B &3.47 &3.04 &1.71 &2.94 &2.43 &3.21 &4.19 &2.24 &3.38 &2.96 \\
    OmniGen2~\cite{wu2025omnigen2}  & 7B &3.57 &3.06 &1.77 &3.74 &3.2 &3.57 &\underline{4.81}& 2.52 &4.68 &3.44\\
    FLUX.1 Kontext [Pro]~\cite{batifol2025flux}  & - &4.25 &4.15 & 2.35 &4.56 &  3.57 &4.26 & 4.57& 3.68 &4.63 & 4.00\\
    Qwen-Image~\cite{wu2025qwen}  & 7B &\underline{4.38} &4.16 &\underline{3.43} &\underline{4.66} & \underline{4.14} &4.38 &\underline{4.81}& 3.82 &\underline{4.69} &\underline{4.27}\\
    GPT Image 1 [High]~\cite{achiam2023gpt} & - & \textbf{4.61} & \textbf{4.33} & 2.9 & 4.35 & 3.66 & \textbf{4.57} & \textbf{4.93} & \textbf{3.96} & \textbf{4.89} & 4.20\\    
     \rowcolor{lightblue} \system  & 7B & 4.31 & 	\underline{4.18} & 	\textbf{3.86}	 & \textbf{4.78}	 & \textbf{4.57}	 & \underline{4.50}	 & 4.69	 & \underline{3.88}	 & 4.00	 & \textbf{4.31}
 \\

    \midrule
    \end{tabular}
    }
    \label{tab:ti2i_sota_cmp}
\end{table*}
\endgroup 
During inference, we follow MaskGIT~\cite{chang2022maskgit} to use the cosine masking schedule and set the default number of generation steps to 50. Moreover, we follow the common practice to employ classifier free guidance scale. Specifically, the guidance scale for text-to-image generation is set to 5.0. As for image editing, we formulate the generation process with classifier free guidance as 
\begin{equation*}
\begin{aligned}
\mathcal{X}_{O} &= \mathcal{P}_{\theta}(\varnothing, \varnothing, \varnothing) \\
&+ s_I \cdot (\mathcal{P}_{\theta} ( \mathcal{X}_{C}^{U}, \varnothing, \mathcal{X}_{C}^{G})- \mathcal{P}_{\theta}(\varnothing, \varnothing, \varnothing)) \\
&+ s_T \cdot ( \mathcal{P}_{\theta} (\mathcal{X}_{C}^{U},\mathcal{T}_C,  \mathcal{X}_{C}^{G}) - \mathcal{P}_{\theta} (\mathcal{X}_{C}^{U},\varnothing,  \mathcal{X}_{C}^{G})),
\end{aligned}
\end{equation*}
where $\mathcal{P}_{\theta}$ represents the parameter of \system, $\varnothing$ denotes empty (drop the condition), $s_T$ refers to the guidance scale of editing instruction and $s_I$ refers to that of the condition image. For evaluation on the ImgEdit benchmark, we set $s_T$ and $s_I$ to 3 and 1.5, respectively.

\begingroup
    \begin{table*}[t]
    \small
    \centering
    \caption{
Comparison with state-of-the-art models on GenEval and DPG-Bench. The best and second-best results are highlighted in \textbf{bold} and \underline{underlined}, respectively. \system achieves the best performance on both benchmarks.
    }
    \footnotesize
    \addtolength{\tabcolsep}{-2pt}
    \resizebox{\linewidth}{!}{%
    \begin{tabular}{lc|ccccc|ccc}
    \toprule
    \multirow{2}{*}{\textbf{Model}} & \multirow{2}{*}{\textbf{\#~Params}} & \multicolumn{5}{c|}{\textbf{GenEval$\uparrow$}} & \multicolumn{3}{c}{\textbf{DPG-Bench$\uparrow$}} \\
\cmidrule(l){3-7}\cmidrule(l){8-10} 
& & \textbf{Two Obj.}  & \textbf{Counting} & \textbf{Position} & \textbf{Color Attri.} & \textbf{Overall}  & \textbf{Attribute}& \textbf{Entity}&  \textbf{Overall}\\
\midrule
\multicolumn{10}{c}{\textit{Text-to-Image Generation Models}}\\
\midrule
DALLE-3\cite{betker2023improving} & - & 0.87 & 0.47 & 0.43 & 0.45 & 0.67 & 88.96 & 89.61 & 83.50 \\
Emu3~\cite{wang2024emu3} & 8B & 0.71 & 0.34 & 0.17 & 0.21 & 0.54 &  88.39 & 86.68 & 80.60\\ 
Infinity~\cite{han2024infinity} & 2B & 0.85 & - & 0.49 & 0.57 & 0.73 & - & 90.76 & 83.46 \\
Lumia-Image-2.0~\cite{qin2025lumina}  & 2.6B & 0.87 & 0.67 & - & 0.62 &0.73 & 90.20 & 91.97 & 87.20 \\
GPT Image 1 [High]~\cite{achiam2023gpt}& - & 0.92 & 0.85 & 0.75 & 0.61 & 0.84  & 89.84 & 88.94  & 85.15\\

\midrule 
\multicolumn{10}{c}{\textit{Unified MLLMs}}\\
\midrule

Janus~\cite{wu2024janus} & 1.3B& 0.68 & 0.30 & 0.46 & 0.42 & 0.61&  87.70  & 87.38 & 79.68 \\
\rowcolor{lightgray} \systemold~\cite{tian2025unigen} & 1.5B & \textbf{0.94}& 0.78 & 0.57 &0.54 &0.78 & \textbf{90.90} & 89.68 & 85.19 \\
Manzano~\cite{li2025manzano} &3B &0.91 & \textbf{0.82} & 0.78 &  \underline{0.71}&  \underline{0.85}& - & - & -\\
TokenFlow-XL~\cite{qu2024tokenflow} & 13B & 0.72 & 0.45 & 0.45 & 0.42 & 0.63 &  81.29 &79.22 & 73.38 \\
Janus-Pro~\cite{chen2025janus} & 7B& 0.89 & 0.59 & \underline{0.79} & 0.66 & 0.80 & 89.40 & 88.90 & 84.19 \\
BAGEL~\cite{deng2025bagel} & 7B MoT & \textbf{0.94} & \underline{0.81} & 0.64 & 0.63 & 0.82 & - & - & -   \\
Show-o2~\cite{xie2025show} & 7B & 0.87& 0.58& 0.52& 0.62& 0.76&  89.96 & \underline{91.78} & \underline{86.14}\\
BLIP3-o~\cite{chen2025blip3}  & 8B & - & -&- & - & 0.84 &- &- & 81.60\\ 
\rowcolor{lightblue} \system & 7B & \underline{0.93}	& 0.80	& \textbf{0.92}	& \textbf{0.81}	&  \textbf{0.89}	& \underline{90.55}	& \textbf{92.64}& \textbf{86.83} \\
\midrule
\end{tabular}
}
\label{tab:t2i_sota_cmp}
\end{table*}
\endgroup
\subsection{Main Results}

We compare \system with state-of-the-art unified MLLMs in \Cref{tab:ti2i_sota_cmp}, \Cref{tab:t2i_sota_cmp} and \Cref{tab:mmu_cmp} and summarize the following findings based on the experimental results.

\noindent \textbf{First, \system obtains competitive performance on image editing benchmarks.} As shown in \Cref{tab:ti2i_sota_cmp}, \system demonstrates state-of-the-art performance on ImgEdit. Without leveraging external diffusion models, \system is leading the benchmark with an overall score significantly outperforming recent models of similar model size such as BAGEL and OmniGen2. Notably, \system even achieves slightly better performance than GPT-Image-1.

\begingroup
    \begin{table*}[t]
    \caption{
Comparison with state-of-the-art models on image understanding benchmarks. $^*$denotes reproduced results. The best and second-best results are highlighted in \textbf{bold} and \underline{underlined}, respectively.
    }
    \centering
    \footnotesize
    \addtolength{\tabcolsep}{2pt} 
    \resizebox{\linewidth}{!}{
    \begin{tabular}{lc|ccccccc}
    \toprule
    \textbf{Model} & \textbf{\#Params} &\textbf{AI2D} &\textbf{GQA} &\textbf{POPE} &\textbf{MMMU} &\textbf{MathVista} & \textbf{ScienceQA}& \textbf{Seedbench} \\
    \cmidrule{1-2}\cmidrule{3-9}
    
    Janus~\cite{wu2024janus} &1.3B&\;\,49.0* & 59.1 & 87.0 & 30.5 & \;\,33.7* & \;\,76.5* & 63.7 \\
    OmniMamba~\cite{zou2025omnimamba} &1.3B&\;\,- & 60.8 & 86.3 & 30.6 & \;\,- & \;\,-  & - \\
    Janus-Pro~\cite {chen2025janus} &1.5B& \;\,63.7* & 59.3 & 86.2 & 36.3 & \;\,36.8* & \;\,75.5* & 68.3 \\
    RecA~\cite {chen2025janus} &1.5B& \ - \, & 58.4 & 83.2 & 35.7  & \;\,- & \;\,-  & 65.3 \\
    ULM-R1~\cite {jiang2025co} &1.5B& \ - \, & - & \textbf{88.9} & \underline{42.3}  & \;\,42.5  & \;\,- & - \\
    Harmon~\cite {wu2025harmonizing} &1.5B& \ - \, & 58.9 & 87.6 & 38.9  & \;\,- & \;\,- & 67.1 \\
    
    \rowcolor{lightgray} \systemold~\cite{tian2025unigen} & 1.5B& 67.4 & 62.3 & 87.8 & 32.3 & 44.6  & \underline{79.4} & \underline{70.8}\\

    UniToken~\cite{jiao2025unitoken} & 7B & 68.7 & - & - & 32.8 & 38.5 & - &69.9 \\
    Show-o2~\cite{xie2025show} & 7B & \textbf{78.6} & \underline{63.1}  & - & \textbf{48.9} & -  & -&69.8  \\
    MUSE-VL~\cite{xie2025muse} & 7B & 69.8  & -  & - & 39.7 & \underline{51.3} & -& 69.1   \\
    MMaDA~\cite{yang2025mmada} & 8B & - & 61.3  & 86.1 & 30.2 & - & - & 64.2  \\    
    \rowcolor{lightblue} \system &7B &  \underline{77.4}	 &\textbf{63.7}	 &\underline{88.3}	 &35.9	 &\textbf{51.9}	 	 &\textbf{86.3} &	\textbf{76.5}
 \\
    \midrule
    \end{tabular}
    }
    \label{tab:mmu_cmp}
\end{table*}
\endgroup

\noindent \textbf{Second, \system achieves promising performance on text-to-image generation benchmarks.} \system yields the final score of 0.89 and 86.83 on GenEval and DPG-Bench, respectively. Compared with \systemold, there's a growth of 0.11 on GenEval and 1.6 on DPG-Bench. \system also beats a range of state-of-the-art unified MLLMs on GenEval, especially on the ``Position'' category. For example, \system significantly outperforms Show-o2, BLIP3-o and BAGEL by 0.13, 0.05 and 0.07 points in overall score. On DPG-Bench, \system largely surpass BLIP3-o by more than 5 points.

\noindent \textbf{Third, \system effectively improves \systemold on understanding benchmarks.} As shown in \Cref{tab:mmu_cmp}, \system significantly improves \systemold across all the benchmarks. We attribute the improvements to three aspects, 1) we scale up the model size to 7B, enhancing the overall capability of the unified MLLM, 2) we increase the resolution of input images and keep the original aspect ratio that are beneficial for maintaining images' native information and 3) we perform understanding-based pre-training, alleviating the mismatch between the training objective for generation and understanding. When compared to other strong unified MLLMs of similar size, \system still demonstrates competitive performance, achieving superior numbers than UniToken, MUSE-VL and MMaDA on most of the benchmarks and on-par results with Show-o2.

\subsection{Ablation Results}

\subsubsection{The impact of Unified RL}

\noindent \textbf{The RL (GRPO) stage significantly improves both image generation and editing tasks.} 
Comparing first row and last row in \Cref{tab:rl_ablation}, we observe a considerable gain introduced by the RL stage, where all three benchmarks are improved with clear margin (from 0.85 to 0.89 in GenEval, from 84.19 to 86.83 in DPG-Bench and from 3.93 to 4.31 in ImgEdit). We also show the comparison qualitatively in \Cref{fig:grpo_impact}. For text-to-image task, \system demonstrates better semantic alignment between text prompts and generated images in diverse scenarios including counting (1st example), position (2nd example) and shape (3rd example). For image editing, we observe that \system achieves finer-grained control over the condition images after GRPO. For example, it successfully makes the ``cat sit up" (1st example) and the ``glass bottle extracted" (last example) which failed before GRPO. Furthermore, we argue that GRPO introduces no performance drop in understanding (See \Cref{sec:und_across_stages}).

\vspace{2pt}
\noindent \textbf{Removing either text-to-image or image editing in RL results in significant performance drop.} When discarding image editing in the RL stage, the image generation benchmarks (GenEval and DPG-Bench) are comparable to \system, but there is large drop on ImgEdit benchmark (row 2 vs. row 4 in \Cref{tab:rl_ablation}). When removing text-to-image in RL training, we observe significant performance degradation on text-to-image generation. Keeping both tasks leads to the best overall performance.

\begin{table}[t!]
\small
\centering
\caption{\textbf{Ablation of Unified RL.} We train \system with different tasks during RL for same steps. T2I stands for text-to-image generation and I-Edit represents image editing. We report the overall score for GenEval, DPG-Bench and ImgEdit benchmarks. We highlight the default setting of \system in {\fboxsep=1pt\colorbox{lightblue}{light blue}}.}
    \addtolength{\tabcolsep}{3pt}
     \small
    \resizebox{0.6\linewidth}{!}
    {\begin{tabular}{cc cccc}
    \toprule
    {\textbf{T2I}} & {\textbf{I-Edit}}  & {\textbf{GenEval}} &  {\textbf{DPG-Bench}}&  {\textbf{ImgEdit}}\\
    \midrule
    & &  0.85 & 84.19 & 3.93\\
    \checkmark & & 0.90 & 86.62 & 4.01 \\
     & \checkmark & 0.85 & 86.39 &4.32\\
      \rowcolor{lightblue}\checkmark & \checkmark & 0.89 &  86.83 & 4.31\\

    \bottomrule
    \end{tabular}} 
\label{tab:rl_ablation}
\end{table}

\subsubsection{The impact of Edit Instruction Alignment}

\noindent \textbf{Edit Instruction Alignment is important prior to RL.} We first evaluate the effect of adding this stage by comparing the results to those in the SFT stage. As shown in \Cref{tab:align_ablation} (row 1 vs. row 2), adding \emph{Edit Instruction Alignment} boost performance of all the three benchmarks even before RL that suggests the advantage of this stage in general.

\vspace{2pt}
\noindent \textbf{The impact of Edit Instruction Alignment is amplified in RL.} As shown in \Cref{tab:align_ablation} (row 3 vs. row 4), adding the \emph{Edit Instruction Alignment} stage is crucial for image editing after RL. Without this stage, \system improves ImgEdit by 0.21 points with RL (row 1 vs. row 3). Benefited from the refined semantic alignment introduced by this stage, RL achieves much larger gain by 0.38 points (row 2 vs. row 4). We also show in \Cref{sec:und_across_stages} that this stage does not sacrifice performance for image understanding.

\begin{table}[t!]
\small
\centering
\caption{\textbf{Ablation of Edit Instruction Alignment.} We report the overall score for GenEval, DPG-Bench and ImgEdit benchmarks. We highlight the default setting of \system in {\fboxsep=1pt\colorbox{lightblue}{light blue}}.}
    \addtolength{\tabcolsep}{-0.5pt}
     \small
    \resizebox{0.6\linewidth}{!}
    {\begin{tabular}{cc cccc}
    \toprule
    {\textbf{Edit Inst.}} & {\textbf{Unified}} &  \multirow{2}{*}{\textbf{GenEval}} &  \multirow{2}{*}{\textbf{DPG-Bench}}&  \multirow{2}{*}{\textbf{ImgEdit}} \\
    {\textbf{alignment}} & {\textbf{RL}} & & & \\
    \midrule
    & & 0.83 & 83.92 & 3.87 \\
    \checkmark & & 0.85 & 84.19 & 3.93\\
     & \checkmark & 0.90& 86.96 & 4.08\\
      \rowcolor{lightblue}\checkmark & \checkmark &0.89	&86.83	&4.31
 \\

    \bottomrule
    \end{tabular}} 
\label{tab:align_ablation}
\end{table}

\section{Conclusion}
\label{sec:conclusion}

We presented \system, a unified MLLM that achieves competitive performance across image understanding, generation and editing tasks. Building upon the \systemold framework, \system enhances the model architecture to extend its capabilities to support image editing and further improves it with our designed \emph{Edit Instruction Alignment} stage. We also proposed a unified RL strategy that jointly optimizes generation and editing via shared reward models, leading to substantial gains in both fidelity and controllability.
Extensive experiments demonstrate that \system achieves state-of-the-art results on a wide range of benchmarks for image understanding, text-to-image generation, and image editing, establishing a strong and extensible baseline to advance future research on unified MLLMs.

\vspace{2pt}
\noindent \textbf{Limitation.} First, \system is not proficient in rendering textual contents (\Cref{fig:failure_case} 1st row). Our model focuses on improving semantic alignment between text instructions and discrete visual tokens and uses only a light-weight visual detokenizer for image reconstruction. This leads to a disadvantage in generating text, which critically relies on preserving fine-grained structural details. We believe that integrating a diffusion-based component into the framework can effectively address this limitation. Second, \system still suffers from visual inconsistency (\Cref{fig:failure_case} last row), a key challenge in image editing tasks. A dedicated reward model is required to enforce visual consistency during RL. We leave this direction for future work.

\vspace{2pt}
\noindent\textbf{Acknowledgment.} We thank Mingze Xu, Peter Fu and Oğuzhan Fatih Kar for their kind help.

\small \bibliographystyle{plainnat} 
\bibliography{main}

\clearpage
\setcounter{page}{1}
\renewcommand{\thetable}{\Alph{table}}
\setcounter{table}{0}
\renewcommand{\thefigure}{\Alph{figure}}
\setcounter{figure}{0}
\appendix

\section{Data Generation Details}
\label{sec:data_generation_details}

\prompt{ht}{Generate Edit Instructions}{
\{\texttt{condition\_image}\} You are a creative assistant specializing in image editing. Your task is to analyze the provided image and generate \{\texttt{num}\} distinct prompts to modify the original images with different categories.

\vspace{2pt}
Be concise yet cover detailed information. Specify the modification with fine-grained visual details related with appearance, action, attributes and location. 

\vspace{5pt}
\textbf{Output Format}:

\vspace{2pt}
Your entire response must be a single block of text. Separate each distinct prompt with a semicolon (;). Do not use bullet points, numbering, or line breaks.

\vspace{5pt}
\textbf{Example of Expected Output}  (3 edit instructions)

\vspace{2pt}
- ``Add a vintage bicycle leaning against the brick wall; Change the color of the car to a deep emerald green; Change the plain brick wall behind the subject to a vibrant, colorful graffiti art wall''

\vspace{2pt}
- ``Remove the briefcase from the man's hand, and change the color of his tie to a deep crimson red; Make the woman turn her head to the left; Extract the pink top shirt the person wears in the image''

\vspace{2pt}
- ``Replace the red car in the foreground with a classic vintage motorcycle; Remove the red car parked by the curb; Transfer the image into a dramatic, high-contrast chiaroscuro painting style''
}{prompt:edit_gen}

We introduce the construction of \textbf{Edit-Align} dataset for the \emph{Edit Instruction Alignment} stage and the \textbf{Edit-RL} dataset for the RL stage. Generally, the data source of Edit-RL is a subset of Edit-Align. For each dataset, we collect the condition images, the edit instructions and the description of desired output images. We describe the data generation pipeline in details as below.

\prompt{ht}{Generation of Desired Output Description}{
You are a precise and objective visual analyst. Your task is to provide a single, standalone, and very short description of an edited image. The description must be 60 words or less.

\vspace{5pt}
You will be given:

\vspace{2pt}
1. An Original Caption describing the original image.

\vspace{2pt}
2. An Editing Prompt describing the change.

\vspace{5pt}
\textbf{Instructuons:}

\vspace{2pt}
1. Be Standalone: Imagine you are describing this final image to someone who has never seen the original. Your description must stand on its own and cannot refer to any previous state or editing process. Avoid using comparative or editing-related words like: `changed`, `now`, `instead of`, `no longer`, `modified`, `edited`, `replaced`, `added`, `removed`, `unlike the original`, `previously`.

\vspace{2pt}
2. Be Faithful: Your description must be a factual account of the final image. Specifically describe the edited element as dictated by the Editing Prompt along with the other key visual elements.

\vspace{2pt}
3. Be Concise: Your entire response must be 60 words or less. Focus only on the most essential information.

\vspace{2pt}
4. Be Faithful: Do not invent details, objects, or attributes that are not visually present.

\vspace{5pt}
Please analyze the following inputs and generate your objective, faithful, and very short description of the final, edited image below.

\vspace{5pt}
Original Caption: \{\texttt{original\_caption\}}

Editing Prompt: \{\texttt{editing\_prompt}\}

}{prompt:output_gen}

\prompt{t}{Assessing Edit and Description Quality}{

\{\texttt{condition\_image}\} You are a meticulous AI Data Quality Analyst. Your task is to score the sample on two criteria: overall description quality and editing quality.

\vspace{5pt}
1. Overall Description Quality Score (0-5): 

\vspace{2pt}
This is a holistic score measuring the quality of the edited image description. A high score requires strength in all of the following areas: (1) The description must accurately reflect the edit requested in the Editing Prompt; (2) It must remain faithful to the unedited parts of the image, without imaginary details, subjective opinions, or hallucinations; (3) It must be a standalone description that does not compare edited elements to their original, unseen state.

\vspace{5pt}
2. Editing Quality Score (0-5): 

\vspace{2pt}
This score evaluates the overall quality of the edit instruction and its outcome based on two factors: (1) the instruction should be plausible given the original image, and (2) it should clearly and unambiguously specify the content to be edited.

\vspace{3pt}
Your response must use the following format:
``Reasoning: \{A brief justification\};\{Overall description quality score\};\{Editing quality score\}''

\vspace{5pt}
Analyze the provided condition image, and the following data. Provide your evaluation in the specified single-line, semicolon-separated format.

\vspace{5pt}
Editing Instruction: \{\texttt{edit\_instruction}\}

Edited Image Caption: \{\texttt{edited\_caption}\}

}{prompt:edit_filter}

\vspace{2pt}
\noindent \textbf{Sourcing condition images.} For Edit-RL dataset, we employ Qwen-Image~\cite{wu2025qwen} to generate 5,000 synthetic images based on text prompts collected from COCO val2017~\cite{lin2014microsoft} and segment anything dataset (SAM)~\cite{kirillov2023segment}. Furthermore, we supplement Edit-Align datasets with around 7,000 images sampled from the BLIP-3o SFT image source.

\vspace{2pt}
\noindent \textbf{Preparing edit instructions.} Given each image in Edit-RL, we leverage Qwen2.5-VL-72B-instruct~\cite{bai2025qwen2} to construct 10 different edit instructions (see \Cref{prompt:edit_gen}). As for image collected from BLIP-3o-SFT in Edit-Align, we generate versatile edit prompts using hand-crafted templates. 

\vspace{2pt}
\noindent \textbf{Generating output descriptions.} As shown in \Cref{prompt:output_gen}, we use a Qwen-2.5-72B-instruct~\cite{yang2024qwen2} model to generate the description for desired output images for training data in Edit-RL. For additional samples in Edit-Align, we use hand-crafted rules to obtain the edited images descriptions. Next, we filter out edit prompts or captions with poor quality by assessing them with Qwen2.5-VL-72B-instruct~\cite{bai2025qwen2} (see \Cref{prompt:edit_filter}). Consequently, we collect 17,663 triplets for Edit-Align and 10,568 triplets for Edit-RL.

\vspace{2pt}
\noindent \textbf{Constructing conversations for instruction alignment.} To perform Post-SFT training with \emph{Edit Instruction Alignment}, we assemble the 17,663 data samples of condition images, edit instructions and output descriptions into instructional conversations, as displayed in \Cref{prompt:edit_align}

\prompt{t}{Conversation in Edit Instruction Alignment}{

\vspace{2pt}
\textbf{User:} 

\vspace{2pt}
\{\texttt{condition\_image}\} Analyze the image and the edit instruction: \{\texttt{edit\_instruction}\}. Write a concise, accurate, and standalone caption describing the final edited image. Focus only on the result of the edit, without referring to or comparing with the original image.

\vspace{5pt}
\textbf{Assistant:} 

\vspace{2pt}
\{\texttt{output\_description}\}

}{prompt:edit_align}

\section{Benchmarks and Evaluation Protocol}
\label{sec:benchmarks}
\noindent\textbf{For image understanding}, we include \textit{(i)} general VQA benchmarks, such as GQA~\cite{hudson2019gqa} and Seedbench \cite{li2023seed},
\textit{(ii)} knowledge-based benchmarks, such as AI2D~\cite{kembhavi2016diagram}, ScienceQA~\cite{lu2022learn}, MMMU~\cite{yue2023mmmu}, and MathVista~\cite{lumathvista},
and \textit{(iii)} hallucination benchmarks, such as POPE~\cite{Li-hallucination-2023}.
We leverage the \texttt{lmms-eval} toolkit to compute the results for the above benchmarks.

\vspace{2pt}
\noindent\textbf{For text-to-image generation benchmarks}, we report results on GenEval~\cite{ghosh2023geneval} and DPG-bench~\cite{hu2024ella} to comprehensively evaluate the semantic alignment between a text prompt and the generated images. We report our results using the official evaluation repository of GenEval and DPG-bench, respectively.

\noindent\textbf{For image editing}, we report results on ImgEdit~\cite{ye2025imgedit} benchmark using the official evaluation repository. All results are evaluated with GPT-4o~\cite{gpt4o}.

\begingroup
\begin{table*}[ht]
    \caption{
        Hyperparameter setup for different training stages of \system. Data ratio refers to the ratio of image understanding data, text understanding data, image generation data and image editing data.
    }
    \centering
    \footnotesize
    \addtolength{\tabcolsep}{12pt} 
    \begin{tabular}{l|cc|c|c}
    \toprule
    \textbf{Hyperparameters} &\textbf{PT} & \textbf{SFT} & \textbf{Edit-Inst-Align} & \textbf{RL} \\
    \hline
    Learning rate & $1e^{-4}$ & $2e^{-5}$ & $1e^{-5}$ &  $3e^{-6}$ \\
    LR scheduler  &  constant & cosine  & cosine  & cosine \\
    Gradient clip & 1.0 & 1.0 & 1.0 & 1.0 \\
    Warm-up steps & 0 & 5000 & 0 & 0\\
    Training steps & 300k & 73k & 0.5k & 1.5k\\
    Batch size  & 576 & 128 & 64 & 32\\
    Data ratio & 2:1:3:- & 8:2:3:3 & 1:-:-:- & -:-:1:1\\
    \midrule
    \end{tabular}
\label{tab:hyperp_train}
\end{table*}
\endgroup

\begin{table*}[htb]
\centering
\footnotesize
\caption{Training data overview of different stages. CC, SA, IMN, T2I-2M, SI, BLIP-3o, GIE stands for CC-3M~\cite{sharma2018conceptual} \& CC-12M~\cite{changpinyo2021conceptual}, SAM-11M~\cite{kirillov2023segment}, ImageNet~\cite{ridnik2021imagenet}, Text-2-Image-2M~\cite{texttoimage2m2024}, ShareGPT-4o-Image~\cite{chen2025sharegpt}, text-to-image data from BLIP-3o-SFT~\cite{chen2025blip3} and GPT-Image-Edit-1.5M~\cite{wang2025gpt}, respectively. }
\addtolength{\tabcolsep}{3pt}
\resizebox{\linewidth}{!}{%
\begin{tabular}{l|cccc}
\toprule
\textbf{Stage} & \textbf{Image Gen Data} & \textbf{Image Edit Data} & \textbf{Und Data} & \textbf{Text-only}\\
\midrule
Pre-training& (CC+SA+IMN) (Recap) & - & (CC+SA+IMN) (Recap) & RefinedWeb\\
Supervised Fine-tuning & BLIP-3o+T2I-2M+SI &SI+GIE &SF-LLaVA1.5 (Image Mixture)~\cite{xu2025slowfast} & RefinedWeb \\
Edit-Inst-Align & -  & -  & Edit-Align& - \\
RL & T2I-R1~\cite{jiang2025t2i} & Edit-RL & - & - \\
\bottomrule
\end{tabular}
}
\label{tab:train_data_overview}
\vspace{-0.1in}
\end{table*}

\section{Training Details}
\label{sec:train_details}
\subsection{Hyper-parameters and Datasets}
An overview of training hyper-parameters is shown in \Cref{tab:hyperp_train}. We also list our training datasets for each training stage in \Cref{tab:train_data_overview}. 

\subsection{Reward Functions}

\vspace{2pt}
\noindent \textbf{Image-text Alignment Model.} We employ DFN5B-CLIP-ViT-H-14~\cite{fang2023data} to model the similarity between a given image and its text prompt. We feed the generated image and the output description into the visual encoder and the text encoder, respectively. Then, we compute the cosine similarity as $\mathbf{R}_C$ between the image and text embeddings for RL training.

\vspace{2pt}
\noindent \textbf{Human Preference Model.}
We use HPSv2~\cite{wu2023human} to evaluate the aesthetic appeal and the alignment between the text description and the generated image. Similarly, we also obtain the cosine similarity between the visual and text features as the reward $\mathbf{R}_H$.

\vspace{2pt}
\noindent \textbf{Semantic Consistency Model.} We leverage the UnifiedReward-7B~\cite{wang2025unified} model to measure the fine-grained consistency (\eg, objects, attributes and relationship) between the text description and the output image. The model outputs a simple reasoning process followed by a final score ranging from 1-5, which is normalized to 0-1 as the reward $\mathbf{R}_U$. 

\vspace{2pt}
\noindent \textbf{Outcome Reward Model.} We take advantage of ORM from ~\cite{guo2025can} to judge whether the generated images correctly represent the given text description. The model is trained to output `Yes' for aligned image-text pairs and yield `No' otherwise. We compute the probability as the reward based on the model’s first-token distribution. Given $p_y$ denoting the probability of first token assigned to `Yes' and $p_n$ indicating the probability for `No', we define the scalar reward as $\mathbf{R}_{O} = \frac{p_y}{p_y + p_n}$.

\vspace{2pt}
\noindent \textbf{Ensemble of Reward Models.} We simply average all reward scores by $\mathbf{R} = mean(\mathbf{R}_C + \mathbf{R}_H + \mathbf{R}_U + \mathbf{R}_O )$. We then perform the advantage computation based on the averaged reward score within each group.

\section{More Results}
\subsection{Understanding Metrics across Training Stages}
\label{sec:und_across_stages}
\vspace{2pt}
Although, both \emph{Edit Instruction Alignment} (Post-SFT) and RL stages are designed to improve image generation and editing, it is also valuable to learn how do they impact \system's image understanding capability. As shown in \Cref{tab:und_across_stages}, there is no performance drop after Post-SFT and RL, suggesting that these stages effectively maintain strong understanding capability while improving image generation and editing.
\begingroup
    \begin{table*}[ht]
    \caption{
    Performance on image understanding benchmarks across different training stages. \emph{Und Avg.} denotes the average score over all the benchmarks.
    }
    \vspace{-5pt}
    \centering
    \footnotesize
    \addtolength{\tabcolsep}{5pt} 
    \resizebox{\linewidth}{!}{
    \begin{tabular}{l|cccccccc}
    \toprule
    \textbf{Model} &\textbf{AI2D} &\textbf{GQA} &\textbf{POPE} &\textbf{MMMU} &\textbf{MathVista} & \textbf{ScienceQA}& \textbf{Seedbench} & \textbf{Und Avg.}\\
    \midrule
    SFT & 77.6	&64.0	&89.2	&35.7	&52.3	&85.9	&76.4 & 68.7
\\
    Edit Inst. Align & 77.6	& 63.8	& 88.6& 	35.7& 	51.7	& 86.3	& 76.5	& 68.6 \\
    RL &  77.4	 &63.7 &88.3	 &35.9	 &51.9	 	 &86.3 &76.5 & 68.6
 \\
    \midrule
    \end{tabular}
    }
    \label{tab:und_across_stages}
\vspace{-8pt}
\end{table*}
\endgroup

\begin{table}[ht]
\small
\centering
\caption{Ablation of condition designs for image editing. We report performance of \system after RL with different sequences of condition in image editing task. $\rightarrow$ denotes the order when concatenating different embeddings. We highlight the default setting of \system in {\fboxsep=1pt\colorbox{lightblue}{light blue}}.}
    \vspace{-0.05in}
    \addtolength{\tabcolsep}{1pt}
     \small
    \resizebox{0.6\linewidth}{!}
    {\begin{tabular}{cccc}
    \toprule
    {\textbf{Condition Input}} &  {\textbf{GenEval}} &  {\textbf{DPG-Bench}}&  {\textbf{ImgEdit}} \\
    \midrule

$\mathcal{X}_{C}^{G}\rightarrow  \mathcal{X}_{C}^{U} \rightarrow \mathcal{T}_C $ & 0.89&	87.47& 4.05\\
 
$\mathcal{X}_{C}^{U}\rightarrow  \mathcal{X}_{C}^{G} \rightarrow \mathcal{T}_C $ & 0.89 & 86.97& 3.98\\

\rowcolor{lightblue} $\mathcal{X}_{C}^{U} \rightarrow \mathcal{T}_C \rightarrow  \mathcal{X}_{C}^{G} $ & 0.89&	86.83& 4.31\\
 \bottomrule
    \end{tabular}} 
\label{tab:edit_cond_ablation}
\vspace{-0.05in}
\end{table}
\subsection{Ablation of Condition Design in Image Editing}

\vspace{2pt}

Inspired by prior works \cite{chen2025sharegpt,wu2025omnigen2}, we utilize semantic ($\mathcal{X}_{C}^{U}$) and low-level ($\mathcal{X}_{C}^{G}$) visual embeddings as well as text embedding $\mathcal{T}_{C}$ as conditions for image editing. When constructing the representation of the conditions, the order of these embeddings is very important. We start with $\mathcal{X}_{C}^{U} \rightarrow \mathcal{T}_{C}$ since this arrangement aligns better with how our model perceives the tokens from visual and text modalities during pre-training, as compared to $\mathcal{T}_{C} \rightarrow \mathcal{X}_{C}^{U} $. Then, there are three options for inserting the $\mathcal{X}_{C}^{G}$ as shown in \Cref{tab:edit_cond_ablation}. From the comparison results, we observe that appending $\mathcal{X}_{C}^{G}$ at the end introduces the best overall result. This is because during pre-training, \system is optimized with $\mathcal{X}_{C}^{U} \rightarrow \mathcal{T}_{C}$ for image understanding and $\mathcal{T}_{C} \rightarrow \mathcal{X}_{C}^{G}$ for image generation so that the design of $\mathcal{X}_{C}^{U} \rightarrow \mathcal{T}_{C} \rightarrow \mathcal{X}_{C}^{G}$ maximally leverages the training momentum.

\begin{figure}[t!]
    \centering
    \includegraphics[width=0.85\linewidth]{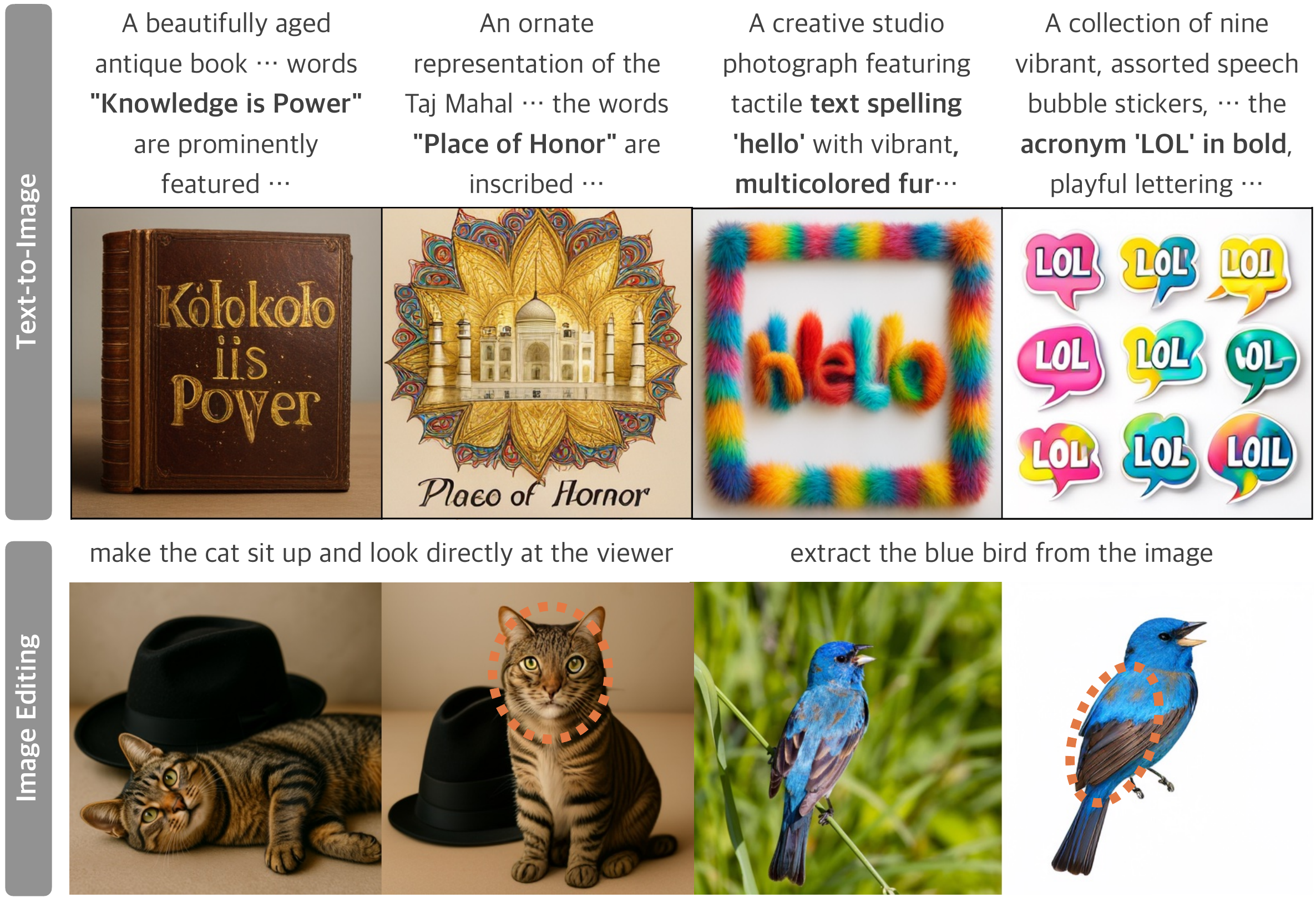}
    \vspace{-10pt}
    \caption{
        Failure cases of \system.}
    \vspace{-10pt}
    \label{fig:failure_case}
\end{figure}

\subsection{The Impact of Different Reward Models}

\begin{table}[ht]
\small
\centering
\caption{\textbf{Ablation of reward models.} We train \system with unified RL using the same training recipe except for the different sets of reward models in each setting. UniR refers to UnifiedReward and DPG stands for the DPG-Bench. We report the overall score for GenEval, DPG-Bench ImgEdit benchmarks. We highlight the default setting of \system in {\fboxsep=1pt\colorbox{lightblue}{light blue}}.}
    \vspace{-0.05in}
    \addtolength{\tabcolsep}{0pt}
     \small
    \resizebox{0.6\linewidth}{!}
    {\begin{tabular}{cccc| ccc}
    \toprule
    \multicolumn{4}{c|}{\textbf{Reward model}} &  \multirow{2}{*}{\textbf{GenEval}} &  \multirow{2}{*}{\textbf{DPG.}}&  \multirow{2}{*}{\textbf{ImgEdit}} \\
    
     {\textbf{HPS}} & {\textbf{CLIP}}& {\textbf{ORM}} & {\textbf{UniR}} \\
    \midrule
    \checkmark & & & & 0.76 & 87.22 & 4.21\\
    \checkmark & \checkmark &  & & 0.85 & 87.13 & 4.28\\
    \checkmark & \checkmark & \checkmark &  & 0.88 & 87.03 & 4.28\\
    \rowcolor{lightblue}\checkmark & \checkmark & \checkmark & \checkmark &  0.89 &  86.83 & 4.31 \\
    \bottomrule
    \end{tabular}} 
\label{tab:rm_ablation}
\vspace{-0.05in}
\end{table}
\vspace{2pt}
Generally, using the ensemble of multiple vision experts strikes a better trade-off between text-to-image and image editing benchmarks. As shown in \Cref{tab:rm_ablation}, using UnifiedReward-7B slightly benefits the ImgEdit benchmark (row 4 vs. row 3). The removal of ORM leads to a drop of 0.03 in the overall score on GenEval (row 2 vs. row 3). Using HPSv2 alone underperforms the combination of HPSv2 and CLIP-H by 0.09 and 0.07 on GenEval and ImgEdit benchmarks, respectively (row 2 vs. row 1).

\subsection{More Visualization}
\label{sec:more_vis}

\vspace{2pt}
\noindent\textbf{Failure cases} of \system in both text-to-image generation and image editing tasks are illustrated in \Cref{fig:failure_case}. In the first row, we present the instances where \system fails to accurately render text characters, as the light-weight discrete detokenizer struggles to control the fine-grained structural details required for text generation. In the second row, we display two examples with visible identity shifts highlighted by the circle, \eg, the changes in cat’s facial fur texture and shape, and the differences in color of the bird's feather. \system needs further improvement to address these limitations.

\end{document}